\documentclass[11pt,a4paper]{article}
\usepackage[hyperref]{acl2021}
\usepackage{times}
\usepackage{latexsym}
\usepackage{graphicx}
\usepackage{multirow}
\usepackage{svg}

\usepackage{microtype}

\aclfinalcopy 
\def\aclpaperid{35} 

\title{AMU-EURANOVA at CASE 2021 Task 1: Assessing the stability of multilingual BERT}

\author{Léo Bouscarrat$^{1,2}$, Antoine Bonnefoy$^{1}$, Cécile Capponi$^{2}$, Carlos Ramisch$^{2}$\\
         $^1$EURA NOVA, Marseille, France \\
         $^2$Aix Marseille Univ, Université de Toulon, CNRS, LIS, Marseille, France \\
         \texttt{\{leo.bouscarrat, antoine.bonnefoy\}@euranova.eu}\\
         \texttt{\{leo.bouscarrat, cecile.capponi, carlos.ramisch\}@lis-lab.fr}}

\date{}

\begin{document}
\maketitle
\begin{abstract}
This paper explains our participation in task 1 of the CASE 2021 shared task. This task is about multilingual event extraction from news. We focused on sub-task 4, event information extraction. This sub-task has a small training dataset and we fine-tuned a multilingual BERT to solve this sub-task. We studied the instability problem on the dataset and tried to mitigate it.
\end{abstract}

\section{Introduction}

Event extraction is becoming more and more important as the number of online news increases. This task consists of extracting events from documents, especially news. An event is defined by a group of entities that give some information about the event. Therefore, the goal of this task is to extract, for each event, a group of entities that define the event, such as the place and time of the event.

This task is related but still different from named entity recognition (NER) as the issue is to group the entities that are related to the same event, and differentiate those related to different events. This difference makes the task harder and also complicates the annotation.

In the case of this shared task, the type of events to extract is protests \citep{Hurriyetoglu+21,Hurriyetoglu+21b}. This shared task is in the continuation of two previous shared tasks at CLEF 2019 \citep{Hurriyetoglu+19b} and AESPEN \citep{Hurriyetoglu+20b}. The first one deals with English event extraction with three sub-tasks: document classification, sentence classification, and event information extraction. The second focuses on event sentence co-reference identification, whose goal is to group sentences related to the same events.

This year, task 1 is composed of the four aforementioned tasks and adds another difficulty: multilinguality. This year's data is available in English, Spanish, and Portuguese. Thus, it is important to note that there is much more data in English than in the other languages. For the document classification sub-task, to test multilingual capabilities, Hindi is available on the testing set only.

We have mainly focused on the last sub-task (event information extraction), but we have also submitted results for the first and second sub-tasks (document and sentence classification). We used multilingual BERT \citep{devlin2019bert}, henceforth M-BERT, which is a model known to obtain near state-of-the-art results on many tasks. It is also supposed to work well for zero-or-few-shot learning on different languages \citep{pires2019multilingual}. We will see the results on these sub-tasks, especially for sub-task 4 where the training set available for Spanish and Portuguese is small.

Thus, one of the issues with transformer-based models such as M-BERT is the instability on small datasets \citep{dodge2020fine, ruder2021lmfine-tuning}. The instability issue is the fact that by changing some random seeds before the learning phase but using the same architecture, data and hyper-parameters the results can have a great variance. We will look at some solutions to mitigate this issue, and how this issue is impacting our results for sub-task 4.\footnote{Our code is available here: \url{https://github.com/euranova/AMU-EURANOVA-CASE-2021}}

\begin{figure*}
    \centering
        \includegraphics[width=\textwidth]{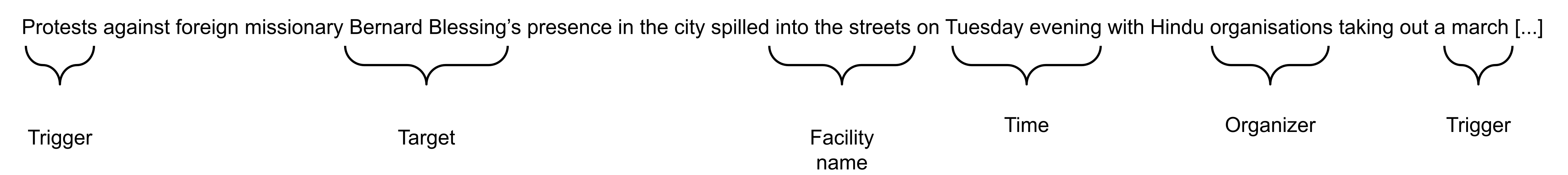}
        \caption{Example of a snippet from sub-task 4.}
        \label{fig:example_subtask4}
\end{figure*}

\section{Tasks and data}

Sub-tasks 1 and 2 can be seen as binary sequence classification, where the goal is to say if a given sequence is part of a specific class. In our case, a classifier must predict whether a document contains information about an event for sub-task 1 or if a sentence contains information about an event for sub-task 2.

Document and sentence classification tasks, sub-tasks 1 and 2,  are not our main research interest. Moreover, the datasets provided for these tasks are less interesting (reasonable amount of training data).

On the other hand, sub-task 4 not only has less training data available but also requires more fine-grained token-based prediction. The goal of sub-task 4 is to extract event information from snippets that contain sentences speaking about the same event. \citet{Hurriyetoglu+19b} have defined that an event has the following information classes (example in Figure~\ref{fig:example_subtask4}):

\begin{itemize}
    \item Time, which indicates when the protest took place,
    \item Facility name, which indicates in which facility the protest took place,
    \item Organizer, which indicates who organized the protest, 
    \item Participant, which indicates who participated in the protest,
    \item Place, which indicates where the protest took place in a more general area than the facility (city, region, ...),
    \item Target, which indicates against whom or what the protest took place,
    \item Trigger, which is a specific word or group of words that indicate that a protest took place (examples: protested, attack, ...), 
\end{itemize}

Thus, not all the snippets contain all the classes, and they can contain several times the same classes. Each information can be composed of one or several adjacent words. Each snippet contains information related to one and only one event.

As the data is already separated into groups of sentences related to the same event, our approach consists of considering a task of named entity recognition with the aforementioned classes.
Multilingual BERT has already been used for multilingual named entity recognition and showed great results compared to state-of-the-art models \citep{hakala-pyysalo-2019-biomedical}.

The data is in BIO format~\citep{ramshaw-marcus-1995-text}, where each word has a B tag or an I tag of a specific class or an O tag. The B tag means beginning and marks the beginning of a new entity. The tag I means inside, which has to be preceded by another I tag or a B tag, and marks that the word is inside an entity but not the first word of the entity. Finally, the O-tag means outside, which means the word is not part of an entity.

\section{System overview}

Our model is based on pre-trained multilingual BERT \citep{devlin2019bert}. This model has been pretrained on multilingual Wikipedia texts. To balance the fact that the data is not equally distributed between all the languages the authors used exponential smoothed weighting to under-sample the most present languages and over-sample the rarest ones. This does not perfectly balance all the languages but it reduces the impact of low-resourced languages. 

The authors of the M-BERT paper shared the weights of a pretrained model that we use to do fine-tuning. Fine-tuning a model consists of taking an already trained model on a specific task and using this model as a starting point of the training for the task of interest. This approach has reached state-of-the-arts in numerous tasks. In the case of M-BERT, the pre-training tasks are Masked Language Modeling (MLM) and Next Sentence Prediction (NSP).

To be able to learn our task, we add a dense layer on top of the outputs of M-BERT and learn it during the fine-tuning. All our models are fine-tuning all the layers of M-BERT.

The implementation is the one from HuggingFace's `transformers' library \citep{wolf-etal-2020-transformers}. To train it on our data, the model is fine-tuned on each sub-task.

\subsection{Sub task 1 and 2}

For sub-tasks 1 and 2, we approach these tasks as binary sequence classification, as the goal is to predict whether or not a document (sub-task 1) or sentence (sub-task 2) contains relevant information about a protest event. Thus the size of the output of the dense layer is 2. We then perform an argmax on these values to predict a class. We use the base parameters in HuggingFace's 'transformers' library. The loss is a cross-entropy, the learning rate is handled by an AdamW optimizer \citep{loshchilov2018decoupled} and the activation function is a gelu \citep{hendrycks2016gaussian}. We use a dropout of 10\% for the fully connected layers inside M-BERT and the attention probabilities.

One of the issues with M-BERT is the limited length of the input, as it can only take 512 tokens, which are tokenized words. M-BERT uses the wordpiece tokenizer \citep{45610}. A token is either a word if the tokenizer knows it, if it does not it will separate it into several sub-tokens which are known. For sub-task 1, as we are working with entire documents, it can be frequent that a document is longer than this limit and has to be broken down into several sub-documents. To retain contexts in each sub-documents we use an overlap of 150 tokens, which means between two sub-documents, they will have 150 tokens in common.  Our method to output a class, in this case, is as follows:

\begin{itemize}
    \item tokenize a document,
    \item if the tokenized document is longer than the 512-tokens limit, create different sub-documents with 150-tokens overlaps between each sub-document,
    \item generate a prediction for each sub-document,
    \item average all the predictions from sub-documents originated from the same document,
    \item take the argmax of the final prediction.
\end{itemize}

\subsection{Sub-task 4}

For sub-task 4, our approach is based on word classification where we predict a class for each word of the documents.

One issue is that as words are tokenized and can be transformed into several sub-tokens we have to choose how to choose the prediction of a multi-token word. Our approach is to take the prediction of the first token composing a word as in \citet{hakala-pyysalo-2019-biomedical}.

We also have to deal with the input size as some documents are longer than the limit. In this case, we separate them into sub-documents with an overlap of 150. Our approach is:

\begin{itemize}
    \item tokenize a document,
    \item if the tokenized document is longer than the 512-tokens limit, create different sub-documents with 150-tokens overlaps between each sub-document,
    \item generate a prediction for each sub-document,
    \item reconstruct the entire document: take the first and second sub-documents, average the prediction for the same tokens (from the overlap), keep the prediction for the others, then use the same process with the obtained document and the next sub-document. As the size of each sequence is 512 and the overlap is only 150, no tokens can be in more than 2 different sequences,
    \item take the argmax of the final prediction for each word.
\end{itemize}

\subsubsection{Soft macro-F1 loss}

We used a soft macro-F1 loss \citep{lipton2014optimal}. This loss is closer than categorical cross-entropy on BIO labels to the metric used to evaluate systems in the shared task. The main issue with F1 is its non-differentiability, so it cannot be used as is but must be modified to become differentiable. The F1 score is based on precision and recall, which in turn are functions of the number of true positives, false positives, and false negatives. These quantities are usually defined as follows:

\[ tp = \sum_{i \in tokens} (pred(i) \times true(i)) \]
\[ fp = \sum_{i \in tokens} (pred(i) \times (1 - true(i))) \]
\[ fn = \sum_{i \in tokens} ((1 - pred(i)) \times true(i)) \]

With:
\begin{itemize}
    \item \textit{tokens}, the list of tokens in a document,
    \item \textit{true(i)}, 0 if the true label of the token i is of the negative class, 1 if the true label is of the positive class
 \item \textit{pred(i)}, 0 if the predicted label of the token i is of the negative class, 1 if the predicted label is of the positive class
\end{itemize}

As we use macro-F1 loss, we compute the F1 score for each class where the positive class is the current class and negative any other class, e.g. if the reference class is B-trigger, then true(i)=1 for B-trigger and \textit{true(i)=0} for all other classes when macro-averaging the F1.

We replace the binary function \textit{pred(i)} by a function outputting the predicted probability of the token i to be of the positive class:
\[ soft\_tp = \sum_{i \in tokens} (proba(i) \times true(i)) \]
\[soft\_fp = \sum_{i \in tokens} (proba(i) \times (1 - true(i))) \]
\[soft\_fn = \sum_{i \in tokens} ((1 - proba(i)) \times true(i)) \]

With \textit{proba(i)} outputting the probability of the token i to be of the positive class, this probability is the predicted probability resulting from the softmax activation of the fine-tuning network.

Then we compute, in a similar fashion as a normal F1, the precision and recall using the soft definitions of the true positive, false positive, and false negative. And finally we compute the F1 score with the given precision and recall. As a loss function is a criterion to be minimized whereas F1 is a score that we would like to maximize, the final loss is $1-F1$.

\subsubsection{Recommendation for improved stability}

A known problem of Transformers-based models is the training instability, especially with small datasets \citep{dodge2020fine, ruder2021lmfine-tuning}.
\citet{dodge2020fine} explain that two elements that have much influence on the stability are the data order and the initialization of the prediction layer, both controlled by pseudo-random numbers generated from a seed. To study the impact of these two elements on the models' stability, we freeze all the randomness on the other parts of the models and change only two different random seeds:
\begin{itemize}
    \item the data order, i.e. the different batches and their order. Between two runs the model will see the same data during each epoch but the batches will be different, as the batches are built beforehand and do not change between epochs,
    \item the initialization of the linear layer used to predict the output of the model. 
\end{itemize}

Another recommendation to work with Transformers-based models and small data made by \citet{mosbach2021on} is to use smaller learning rates but compensating with more epochs. We have taken this into account during the hyper-parameter search. 

\citet{ruder2021lmfine-tuning} recommend using behavioral fine-tuning to reduce fine-tuning instabilities. It is supposed to be especially helpful to have a better initialization of the final prediction layer. It has also already been used on named entity recognition tasks \citep{broscheit-2019-investigating} and has shown that it has improved results for a task with a very small training dataset.
Thus, to do so, we need a task with the same number of classes, but much larger training datasets. As we did not find such a task, we decided to fine-tune our model on at least the different languages we are working with, English, Spanish and Portuguese. We used named entity recognition datasets and kept only three classes in common in all the datasets: person, organization, and location. These three types of entities can be found in the shared task.

To perform this test, the training has been done like that:

\begin{itemize}
    \item the first fine-tuning is done on the concatenation of NER datasets in different languages, once the training is finished we save all the weights of the model,
    \item we load the weights of the previous model, except for the weights of the final prediction layer which are randomized with a given seed,
    \item we train the model on the dataset of the shared task.
\end{itemize}

\section{Experimental setup}

\subsection{Data}

The dataset of the shared task is based on articles from different newspapers in different languages. More information about this dataset can be found in \cite{Hurriyetoglu+21}

For the final submissions of sub-tasks 1, 2, and 4 we divided the dataset given for training purposes into two parts with 80\% for training and 20\% for evaluation during the system training phase. We then predicted the data given for testing purposes during the shared task evaluation phase. The quantity of data for each sub-task and language can be found in Table~\ref{table:trainingdata}. We can note that the majority of the data is in English. Spanish and Portuguese are only a small part of the dataset.

\begin{table}
\begin{center}
\begin{tabular}{ | c | c | c | c | }
 \hline
 Sub-task & English & Spanish & Portuguese \\
 \hline
 Sub-task 1 & 9,324 & 1,000 & 1,487 \\  
 Sub-task 2 & 22,825 & 2,741 &  1,182 \\
 Sub-task 4  & 808 &  33 & 30 \\
 \hline
 
\end{tabular}
\caption{Number of elements for each sub-task for each language in the data given for training purposes. Documents for sub-task 1, sentences for sub-task 2, snippet (group of sentences about one event) for sub-task 4.}
\label{table:trainingdata}
\end{center}
\end{table}

For all the experiments made on sub-task 4, we divided the dataset given for training purposes into three parts with 60\% for training, 20\% for evaluating and 20\% for testing.

To be able to do our approach of behavioral fine-tuning, we needed some Named Entity Recognition datasets in English, Spanish and Portuguese. For English we used the CoNLL 2003 dataset \citep{tjong-kim-sang-de-meulder-2003-introduction}, for Spanish the Spanish part of the CoNLL 2002 dataset \citep{tjong-kim-sang-2002-introduction} and for Portuguese the HAREM dataset \citep{santos2006harem}. Each of these datasets had already three different splits for training, development and test. Information about their size can be found in Table~\ref{table:behavioraldata}.

\begin{table}
\begin{center}
\begin{tabular}{ | c | c | c | c | }
 \hline
 Dataset & Train & Eval & Test \\
 \hline
 CoNLL 2003 & 14,041 & 3,250 & 3,453 \\  
 CoNLL 2002 & 8,324 & 1,916 & 1,518 \\
 HAREM & 121 & 8 & 128 \\
 \hline
\end{tabular}
\caption{Number of elements for each dataset used in the behavioral fine-tuning in each split.}
\label{table:behavioraldata}
\end{center}
\end{table}

The dataset for Portuguese is pretty small compared to the two others, but the impact of the size can be interesting to study.

\begin{figure*}
    \centering
        \includegraphics[width=\textwidth]{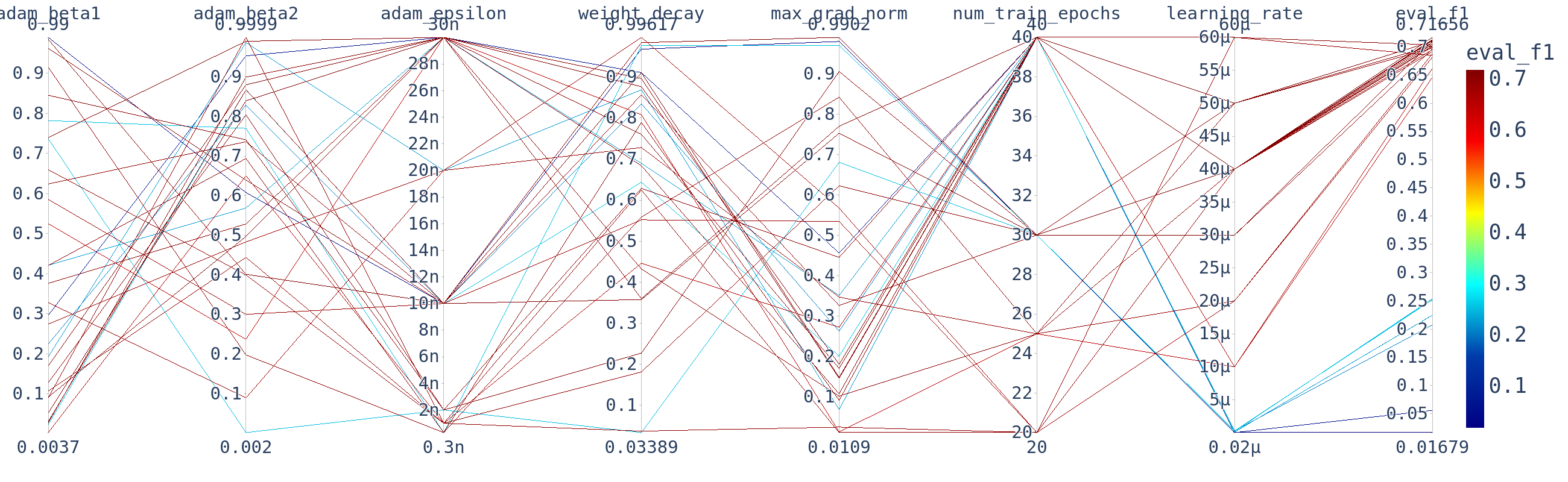}
        \includegraphics[width=\textwidth]{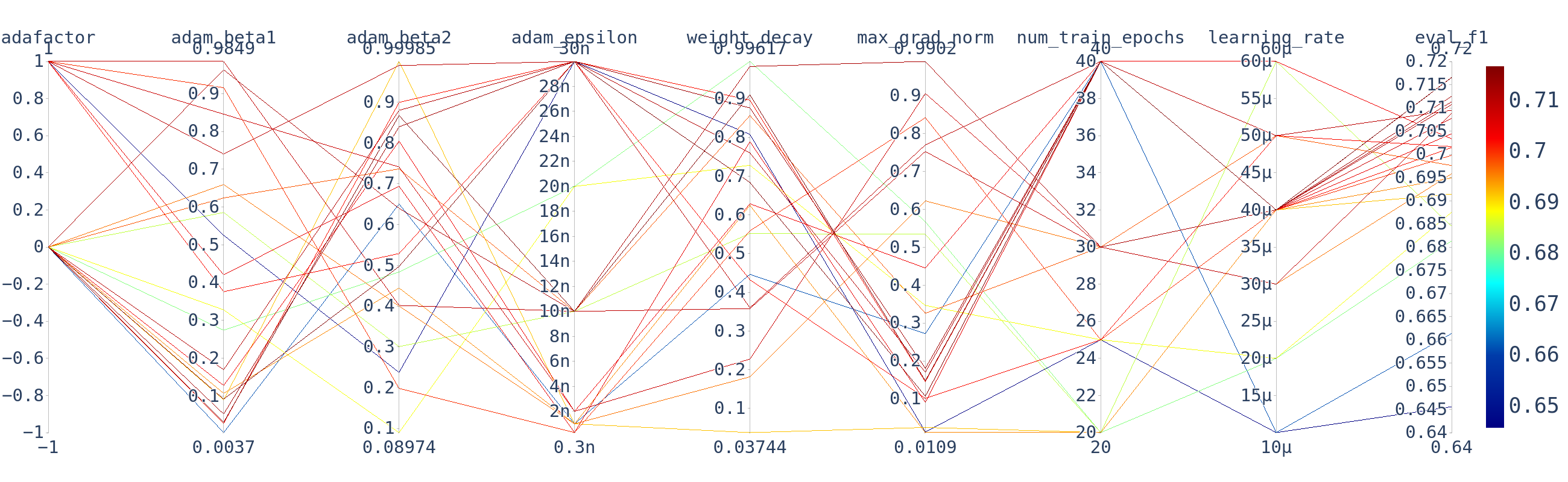}
        \caption{(Top) Parallel coordinates plot of the 30 experiments on sub-task 4 during the hyper-parameter search in function of the value of the hyper-parameters and the value of the F1 on the evaluation set. Each line represents an experiment, and each column a specific hyper-parameter, except the last which is the value of the metric. (Bottom) Same plot with the worst results removed to have a better view of the best results.}
        \label{fig:hyper-parameter_search}
\end{figure*}

\subsection{Hyper-parameter search}

For sub-task 4, we did a hyper-parameter search to optimize the results. We used Ray Tune \citep{liaw2018tune} and the HyperOpt algorithm \citet{bergstra2013making}. We launched 30 different trainings, all the information about the search space and the hyper-parameters can be found in~\ref{subsection:hyper-parameter}. The goal is to optimize the macro-F1 on the evaluation set.

Our goal was to find a set of hyper-parameters that performs well to use always the same in the following experiments. We also wanted to evaluate the impacts of the hyper-parameters on the training.

\subsection{Behavioral fine-tuning}

For the first part of the behavioral fine-tuning, we trained an M-BERT model on the three NER datasets for one epoch. We only learn for one epoch for timing issues, as the learning on this datasets takes several hours.
We then fine-tune the resulting models with the best set of hyper-parameters found with the hyper-parameter search.

\subsection{Stability}

To study the stability of the model and the impact of behavioral fine-tuning we made 6 sets of experiments with 20 experiments in each set:

\begin{itemize}
    \item normal fine-tuning with random data order and frozen initialization of final layer,
    \item normal fine-tuning with frozen data order and random initialization of final layer,
    \item normal fine-tuning with random data order and random initialization of final layer,
\item behavioral fine-tuning with random data order and frozen initialization of final layer,
    \item behavioral fine-tuning with frozen data order and random initialization of final layer,
    \item behavioral fine-tuning with random data order and random initialization of final layer,
\end{itemize}

Once again it is important to note that what we called behavioral fine-tuning is different from behavioral fine-tuning as proposed by \citet{ruder2021lmfine-tuning}, as we reset the final layer. Only the weights of all the layers of M-BERT are modified.

For each set of experiments we will look at the average of the macro-F1, as implemented in \citet{seqeval}, and the standard deviation of the macro-F1 on the training dataset, on the evaluation dataset, and on three different test datasets, one for each language.
Thus we will be able to assess the importance of the instability, if our approach to behavioral fine-tuning helps to mitigate it and if it has similar results across the languages.

We can also note that in our implementation the batches are not randomized. They are built once before the learning phase and do not change, neither in content nor order of passage, between each epoch.

\section{Results}

\subsection{Hyper-parameter search}

The results of the hyper-parameter search can be seen in Figure~\ref{fig:hyper-parameter_search}. On the top pictures which represent the 30 experiments, we can see that a specific hyper-parameter seems to impact the worst results (in blue). This parameter is the learning rate, we can see it in the red box on the top image, all the blue lines are at the bottom, which means these experiments had a small learning rate. It seems that we obtain the best results with a learning rate around 5e-05 (0.00005), lower than 1e-06 seems to give bad results.

\begin{table*}[ht]
\begin{center}
\begin{tabular}{|c|c|c|c|c|c|c|c|}
    \hline
    & Data & Init layer & Train & Eval & Test EN & Test ES & Test PT \\
    \hline
    \multirow{3}{*}{N} & Rand & Fix & 86.11 (1.08) & 69.34 (1.01) & 71.80 (.85) & 54.33 (3.43) & 73.14 (\textbf{1.96}) \\
    & Fix & Rand & \textbf{86.88} (\textbf{.53}) & \textbf{70.03} (.63) & 71.68 (\textbf{.53}) & 55.02 (3.28) & 74.51 (2.41) \\
    & Rand & Rand & 86.63 (1.08) & 69.56 (.97) & \textbf{71.94} (.72) &54.73 (3.44) &74.08 (3.37) \\
    \hline
    \multirow{3}{*}{B} & Rand & Fix & 85.79 (.97) & 69.32 (1.00) & 71.60 (.54) &54.69 (2.99) &74.01 (2.92) \\
    & Fix & Rand & 86.20 (.55) & 69.57 (\textbf{.51}) & 71.80 (.58) & 53.97 (3.90) & 74.50 (2.67) \\
    & Rand & Rand & 86.11 (.87) & 69.40 (.80) &71.85 (.73) &\textbf{55.51} (\textbf{2.82}) &\textbf{74.97} (2.66) \\
    \hline
\end{tabular}
\caption{Average macro-F1 score, higher is better (standard deviation, lower is better) of the 20 experiments with the specified setup. N means normal fine-tuning and B behavioral fine-tuning. Data means data order and Init layer means initialization of the final layer. Rand means random, and fix refers to frozen.}
\label{table:stability}
\end{center}
\end{table*}

\begin{table*}
 \begin{center}
    \begin{tabular}{|l|l|l|l|l|}
        \hline
                   & English & Spanish & Portuguese & Hindi  \\ \hline
        Sub-task 1 & 53.46 (84.55) & 46.47 (77.27) & 46.47 (84.00)  & 29.66 (78.77) \\ \hline
        Sub-task 2 & 75.64 (85.32) & 76.39 (88.61) & 81.61 (88.47)  & /      \\ \hline
        Sub-task 4 & 69.96 (78.11) & 56.64 (66.20)  & 61.87 (73.24)    & /      \\ \hline
    \end{tabular}
 \caption{Score of our final submissions for each sub-task, in parenthesis the score achieved by the best scoring team on each sub-task.}
 \label{table:resultfinal}
 \end{center}
\end{table*}
 
We can then focus on the bottom picture, with the same type of plot but with the worst results removed. Another hyper-parameter that seems to have an impact is the number of training epochs, 40 seems better than 20. We use a high number of epochs as recommended by \citet{mosbach2021on} to limit the instability. Beyond the learning rate and number of epochs, it is then hard to find impactful hyper-parameters.

Finally, the set of hyper-parameters that has been selected is:
\begin{itemize}
    \item Adafactor: True
    \item Number of training epochs: 40
    \item Adam beta 2: 0.99
    \item Adam beta 1: 0.74
    \item Maximum gradient norm: 0.17
    \item Adam epsilon: 3e-08
    \item Learning rate: 5e-05
    \item Weight decay: 0.36
\end{itemize}

For the stability experiments, the number of training epochs have been reduced to 20 for speed purposes. For the first part of the behavioral fine-tuning, the learning rate has been set to 1e-05 as more data were available.

\subsection{Behavioral fine-tuning}

The results on the test dataset of each model after one epoch of training can be found in Table~\ref{table:resultbehavioral}.
\begin{table}[!h]
\begin{center}
\begin{tabular}{| c | c |}
 \hline
 Dataset & Test macro-F1 \\
 \hline
 CoNLL 2003 & 89.8 \\  
 CoNLL 2002 & 86.1 \\
 HAREM & 76.1 \\
 \hline
\end{tabular}
\caption{Macro-F1 score of the NER task on the test split of each dataset used in behavioral fine-tuning after training the base M-BERT for 1 epoch.}
\label{table:resultbehavioral}
\end{center}
\end{table}

We could not compare to state-of-the-art NER models on these three datasets as we do not take all the classes (classes such as MISC were removed before the learning phase). The metrics used on these datasets are not by classes, so the comparison cannot be made. However, the results are already much better than what a random classifier would output, thus the weights of the models should already be better than the weights of the base model.

\subsection{Stability}

The results of the different sets of experiments can be found in Table~\ref{table:stability}.
First, we can see that the difference between behavioral fine-tuning and normal fine-tuning is not important enough to say one is better than the other.
We can also note that the standard deviation is small for English, but not negligible for Spanish and Portuguese.

\subsection{Final submission}

The results of the final submissions can be found in Table~\ref{table:resultfinal}. We can see that our results are lower than the best results, especially for sub-task 1 with a difference of between 30 to 50 macro-F1 score depending on the language, whereas for sub-tasks 2 and 4 the difference is close to 10 macro-F1 score for all the languages.

\section{Conclusion}

\subsection{Sub-task 1 and 2}

As we can see in Table~\ref{table:resultfinal}, our final results for sub-task 1 are much lower than the best results, but for sub-task 2 the difference is smaller. This is interesting as the tasks are pretty similar, thus expected the difference between our results and the best results to be of the same magnitude.

One explanation could be our approach to handle documents longer than the input of M-BERT. We have chosen to take the average of the sub-documents, but if one part of a document contains an event the entire document does too. We may have better results looking if one sub-document at least is considered as having an event.

It is then hard to compare to other models as we have chosen to use one model for all the languages and we do not know the other approaches.

\subsection{Sub-task 4}

For sub-task 4 we have interesting results for all the languages, even for Spanish and Portuguese, as we were not sure that we could learn this task in a supervised fashion with the amount of data available. In a further study, we could compare our results with results obtained by fine-tuning monolingual models, where we fine-tune one model for each language with only the data of one language. This could show the impact of having data if using a multilingual model instead of several monolingual models improves or not the results. We do not expect good results for Spanish and Portuguese as the training dataset is pretty limited. The results seem to comfort the claim of \citep{pires2019multilingual} that M-BERT works well for few-shot learning on other languages.

The other question for sub-task 4 was about instability. In Table~\ref{table:stability} we can see that the instability is way more pronounced for Spanish and Portuguese. It seems logical as we have fewer data available in Spanish and Portuguese than in English. The standard deviation for Spanish and Portuguese is large and can have a real impact on the final results. Finding good seeds could help to improve the results for Spanish and Portuguese.

Furthermore, our approach of behavioral fine-tuning did not help to reduce the instabilities. It was expected that one of the sources of the instability is the initialization of the prediction, and in our approach, the initialization of this layer is still random. In our approach, we only fine-tune the weights of M-BERT. This does not seem to work and reinforces the advice of \citet{ruder2021lmfine-tuning} that using behavioral fine-tuning is more useful for having a good initialization of the final prediction layer.

On the two sources of randomness we studied, data order seems the most impactful for English, where we have more data. Nonetheless, for Spanish and Portuguese, the two sources have a large impact. In a further study, we could see how the quantity of data helps to decrease the impact of these sources of instabilities.

For the final submissions, the macro-F1 score for English and Portuguese is beneath the average macro-F1 score we found during our development phases. This could be due to bad seeds for randomness or because the splits are different. We did not try to find the best-performing seeds for the final submissions.

\section*{Acknowledgments}

We thank Damien Fourrure, Arnaud Jacques, Guillaume Stempfel and our anonymous reviewers for their helpful comments. 


\bibliographystyle{acl_natbib}
\bibliography{papers}

\appendix

\section{Appendix}

\subsection{Hyper-parameter search}
\label{subsection:hyper-parameter}

The space search for our hyper-parameter search was:

\begin{itemize}
    \item Number of training epochs: value in [20, 25, 30, 40],
    \item Weight decay: uniform distribution between 0.001 and 1,
    \item Learning rate: value in [1e-5, 2e-5, 3e-5, 4e-5, 5e-5, 6e-5, 2e-7, 1e-7, 3e-7, 2e-8],
    \item Adafactor: value in "True", "False",
    \item Adam beta 1: uniform distribution between 0 and 1,
    \item Adam beta 2: uniform distribution between 0 and 1,
    \item Epsilon: value in [1e-8, 2e-8, 3e-8, 1e-9, 2e-9, 3e-10],
    \item Maximum gradient norm: uniform distribution between 0 and 1.
\end{itemize}

For the HyperOpt algorithm we used two set of hyper-parameters to help finding a good sub-space. We maximized the macro-F1 on the evaluation dataset, and set the number of initial points before starting the algorithm to 5.
\end{document}